\documentclass[11pt]{article}

\usepackage[final]{acl}

\usepackage{times}
\usepackage{latexsym}

\usepackage[T1]{fontenc}

\usepackage[utf8]{inputenc}

\usepackage{microtype}

\usepackage{inconsolata}

\usepackage{graphicx}

\usepackage{booktabs}
\usepackage{amsmath}
\usepackage{amssymb}
\usepackage{tikz}
\usetikzlibrary{arrows.meta,positioning,calc,fit,backgrounds}
\usepackage{hyperref}
\usepackage{url}

\title{Augmenting Pretrained Language Models with N-gram Embeddings through Incremental Training}
\author{
 \textbf{Shaohua Hu\textsuperscript{1,$\dagger$}},
 \textbf{Xusheng Gu\textsuperscript{2,$\dagger$}},
 \textbf{Haofei Yang\textsuperscript{2}},
 \textbf{Minghao Shao\textsuperscript{2}},
 \textbf{Nanhui Zhao\textsuperscript{2}},
\\
 \textbf{Caoyun Fan\textsuperscript{2,*}},
 \textbf{Fangming Li\textsuperscript{2,*}},
\\
\\
 \textsuperscript{1}MoE Key Lab of Artificial Intelligence, AI Institute, Shanghai Jiao Tong University, Shanghai, China\\
 \textsuperscript{2}ICT AI Competence Center, Huawei Technologies Co., Ltd, Shanghai, China
\\
 \small{
   \textsuperscript{$\dagger$}These authors contributed equally.
 }
\\
 \small{
   \textsuperscript{*}\textbf{Corresponding Author}
 }
}
\begin{document}
\maketitle
\begin{abstract}

As a new scaling technique for large language models, the n-gram embedding improves the
base ability of a model while adding almost no inference cost.
Mainstream research realizes it through joint training of the n-gram embedding with the
backbone, which severely restricts the settings in which an n-gram embedding can be used.
We therefore propose training an n-gram embedding onto an existing model through
incremental training, to extend where the technique can deliver value.
Specifically, we explore several settings of the n-gram embedding under incremental
training---the learning rate of the embedding, the vocabulary size, and the injection
position---and train an n-gram embedding onto the existing model under the best
configuration.
Experiments show that the best configuration under incremental training does not fully
coincide with that under pretraining; moreover, the incrementally trained knowledge is
written predominantly into the n-gram embedding, while the general ability of the
backbone is retained throughout this training process.
\end{abstract}

\section{Introduction}

The n-gram embedding represents n adjacent tokens as a whole, and is a promising model
scaling technique.
It allows local context to be stored and retrieved directly, rather than being
reconstructed by the backbone.
LongCat-Flash-Lite adds an n-gram embedding branch at the embedding layer of the backbone
\citep{liu2026scaling};
Engram fuses n-gram embeddings into the hidden states of the shallow and middle layers
\citep{cheng2026conditional};
Qwen3.8-Next finds experimentally that placing the n-gram embedding at the second layer
yields the best average performance \citep{qwen2026next}.
However, in all of these works the n-gram embedding is pretrained jointly with the
backbone, and what the n-gram embedding can do under incremental training remains
unstudied.

\begin{figure*}[!ht]
  \centering
  \includegraphics[width=\textwidth]{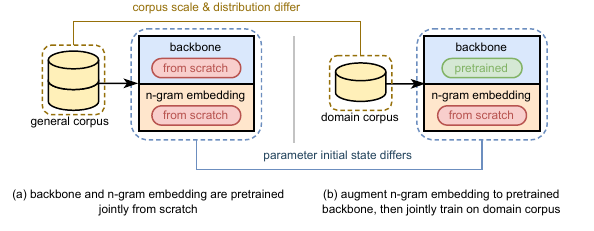}
  \caption{(a) Existing works introduce the n-gram embedding at the pretraining stage
  and jointly optimize it with the backbone from scratch;
  (b) this paper considers incrementally attaching an n-gram embedding to a pretrained
  model that originally lacks the module, to learn domain knowledge.
  The two scenarios differ in the scale of the corpus and the initial state of the
  parameters.}
  \label{fig:overview}
\end{figure*}

The benefit of the n-gram embedding is therefore confined to the pretraining stage.
The goal of this paper is to train an n-gram embedding onto pretrained models that
originally lack the module (Figure~\ref{fig:overview}).

Unlike joint pretraining, incremental training performs domain enhancement with
limited domain data, and only the parameters of the n-gram
embedding module start from their initial state. To find out how the two scenarios
differ, we examine three factors of the n-gram embedding: the learning rate, the
vocabulary size, and the position at which it is injected.
We find that the configuration of the n-gram embedding differs between the two
scenarios in the following ways:
the n-gram embedding needs a larger learning rate to be trained effectively;
a smaller vocabulary than in joint pretraining suffices to learn domain knowledge effectively;
and while among the jointly pretrained works Engram \citep{cheng2026conditional} chooses
shallow and middle layers and Qwen3.8-Next \citep{qwen2026next} chooses shallow layers
only, we find that covering both shallow and middle
layers works best under incremental training.

We validate the approach on a question-answering task in the telecom domain, where the
data contains domain-specific terms, system versions, and configuration parameters.
This information appears as multiple consecutive tokens, and is exactly what the n-gram
embedding can store and retrieve.
After training the n-gram embedding incrementally, model performance improves by
$10.46$ points on average over the base model.
To further examine the role of the n-gram embedding in incremental training, we unload it
and keep the backbone alone.
Once unloaded, the domain ability of the model drops markedly while its general ability
is retained, showing that domain knowledge is written predominantly into the n-gram
embedding and that the general ability of the backbone is left untouched.

The contributions of this paper are as follows:
\begin{enumerate}
  \item We extend the n-gram embedding from joint pretraining to the incremental training
  setting, augmenting pretrained models with n-gram embeddings.
  \item We analyze the three factors of the n-gram embedding under incremental
  training---the learning rate, the vocabulary size, and the injection position---and find
  that the optimal choices differ from those under joint pretraining.
  \item We validate the approach on a question-answering task in the telecom domain:
  training the n-gram embedding incrementally yields significant performance gains;
  moreover, it does not harm the general ability of the backbone, and the domain ability
  is trained predominantly into the n-gram embedding.
\end{enumerate}

\section{Related Work}
\textbf{Word embeddings.} A word embedding maps discrete tokens to vectors, so that the
similarity between tokens can be measured directly; it is the fundamental component with
which a language model quantifies text.
Early word embeddings take a single token as the retrieval key
\citep{bengio2003neural,mikolov2013efficient};
fastText \citep{bojanowski2017fasttext} represents a word as the sum of its character
n-gram vectors, extending the retrieval key from a single token to n-grams.
Owing to the static-lookup property of word embeddings, recent work exploits them to
scale model capacity without significantly increasing computation:
Gemma 3n's per-layer embeddings (PLE) \citep{gemma3n_googleai} and STEM
\citep{sadhukhan2026stem} remove the constraint that embeddings sit only at the input;
N-Grammer \citep{roy2022n}, the Over-tokenized Transformer \citep{huang2025overtok}, and
SCONE \citep{yu2026scaling} extend the key from a single token to n-grams;
Engram \citep{cheng2026conditional}, LongCat-Flash-Lite \citep{liu2026scaling}, and
Qwen3.8-Next \citep{qwen2026next} explore the performance of combining the two scaling
routes, n-gram embeddings and MoE.
This paper further explores using the n-gram embedding to extend pretrained models that
lack it.

\textbf{Incremental training.} The most direct form of incremental training is to train
the model directly on domain data, which may however cause overfitting and catastrophic
forgetting.
A large body of work therefore freezes the backbone and trains only newly introduced
modules, learning new tasks while leaving the backbone parameters untouched as far as
possible.
Adapter tuning \citep{houlsby2019adapter} adapts to downstream tasks efficiently by
inserting small modules between Transformer layers; the technique has since been applied
to CLIP and vision transformers for downstream transfer
\citep{gao2021clipadapter,chen2022vision};
LoRA \citep{hu2021lora} adapts to downstream tasks with low-rank matrices in place of
backbone updates;
ControlNet \citep{zhang2023controlnet} copies blocks of an existing network as a parallel
branch, allowing text-to-image generation to additionally take reference images as input;
Flamingo \citep{alayrac2022flamingo} combines pretrained vision and language
models through attention layers, adapting to multimodal tasks via incremental training.
These works provide valuable references for our work.

\section{Preliminary}
Let $\theta$ denote the parameters of the backbone, and $\phi$ the parameters of the
n-gram embedding branch to be augmented.
We adopt Engram \citep{cheng2026conditional} as the implementation of the n-gram
embedding.
Its forward computation involves two key steps, retrieval and injection.
The retrieval step takes the $n$ tokens ending at position $t$,
$\{x_{t-n+1},\ldots,x_t\}$, as the key, and looks up the table to obtain the n-gram
embedding $\mathbf{e}_t$.
The injection step additively injects $\mathbf{e}_t$ into the hidden state
$\mathbf{h}_{l,t}$ of each injection layer $l$, yielding the updated hidden state
\begin{equation}
  \label{eq:inject}
  \mathbf{h}'_{l,t} = \mathbf{h}_{l,t} + \mathbf{y}_{l,t},
\end{equation}
where the injection vector $\mathbf{y}_{l,t}$ is the n-gram embedding $\mathbf{e}_t$
projected by $\mathbf{W}_V$ and weighted by a context-aware gate $\alpha_{l,t}$:
\begin{equation}
  \label{eq:inject-vec}
  \mathbf{y}_{l,t} = \alpha_{l,t}\mathbf{W}_V \mathbf{e}_t,
\end{equation}
with $\mathbf{W}_V$ the projection matrix.
The gate decides the merging strength according to the relevance between the current
hidden state and $\mathbf{e}_t$, preventing irrelevant n-gram embeddings, as well as
erroneous embeddings caused by hash collisions, from disturbing the output of the
backbone:
\begin{equation}
  \label{eq:gate}
  \alpha_{l,t} = \sigma\left(\frac{\text{RMSNorm}(\mathbf{h}_{l,t})^\top \text{RMSNorm}(\mathbf{W}_K \mathbf{e}_t)}{\sqrt{d}}\right),
\end{equation}
where $\sigma$ is the activation function, $d$ is the hidden-state dimension, and
$\mathbf{W}_K$ is the projection matrix.
The training of the n-gram embedding branch thus involves two goals: the embedding table
should store the correct vector for each n-gram, and the gate should activate in
appropriate contexts, so that the injected vector perturbs the backbone effectively.
During incremental training, $\theta$ and $\phi$ are jointly optimized on the domain
corpus $\mathcal{D}$ under the standard language modeling objective.
\begin{figure*}[!ht]
  \centering
  \includegraphics[width=\textwidth]{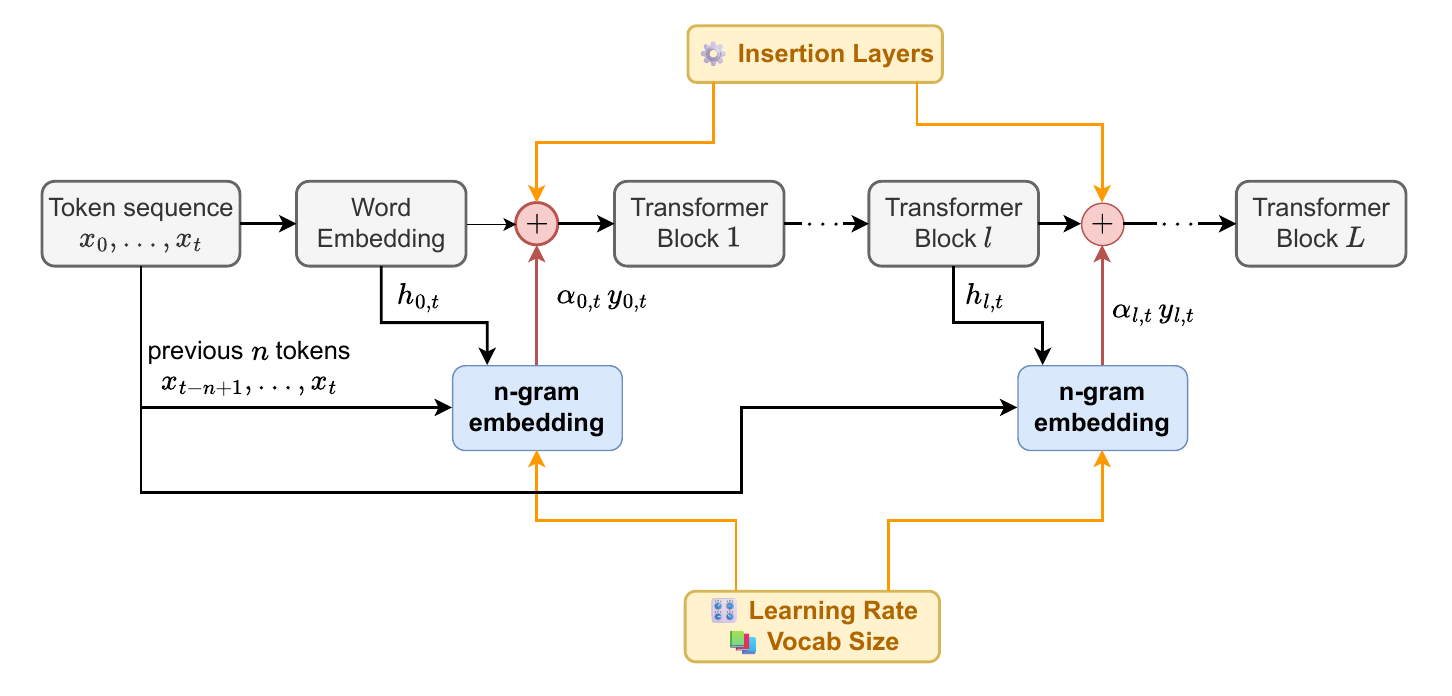}
  \caption{The side-branch structure of Engram and the three key configurations
  during augmentation and incremental training (yellow boxes).
  The n-gram embedding branch leaves the backbone's computation graph untouched:
  $\mathbf{e}_t$ is retrieved with the $n$ tokens ending at position $t$ as the key and,
  weighted by the gate $\alpha_{l,t}$, additively merged into the residual stream
  at the insertion layers.
  The insertion layers are chosen from $[0,\ldots,L]$ and multiple layers may be used;
  the vocabulary size determines the number of slots in the n-gram embedding table;
  both are fixed once the model is augmented.
  The learning rate is set separately for the backbone and the n-gram embedding,
  forming the dual-learning-rate scheme.}
  \label{fig:engram}
\end{figure*}

\section{Method}
\label{sec:method}
\label{sec:method-setup}
To make the n-gram embedding effective under incremental training, we test three
factors---the learning rate, the vocabulary size, and the injection position.
Determining the three factors requires repeated training and evaluation over multiple
configurations, which is costly; limited by our compute resources, we adopt the
small-scale Qwen3-0.6B as the backbone, train it on a question-answering task in the
telecom domain, and determine the optimal setting by the model's performance on the
test set.
Before the optimal setting is determined, the n-gram learning rate is set to
$1\times 10^{-6}$, the vocabulary size equals that of the tokenizer, $152\mathrm{k}$, and
the injection positions are the layer 0 (embedding layer) and the layer 14 (middle layer).
The backbone learning rate is always $1\times 10^{-6}$.

\subsection{Learning rate of the n-gram embedding}
\label{sec:method-dual-lr}
The learning rate decides how much the n-gram embedding is updated during incremental
training, and it is the precondition for the n-gram embedding to work:
if the n-gram embedding is not updated effectively, it cannot perturb the output of the
backbone, and augmenting degenerates into ordinary incremental training of the backbone
alone.
In joint pretraining the backbone and the n-gram embedding are initialized at the same
time, whereas under incremental training only the n-gram embedding starts from its
initialization: it must learn the entire embedding table from scratch, and at each step
only the retrieved slots are updated.
This difference between the two scenarios means that whether the learning-rate setting of
joint pretraining carries over must be examined first.
We first examine the setting that assigns the same learning rate,
$1\times 10^{-6}$, to the backbone and the n-gram embedding.

\begin{figure*}[h]
  \centering
  \includegraphics[width=\textwidth]{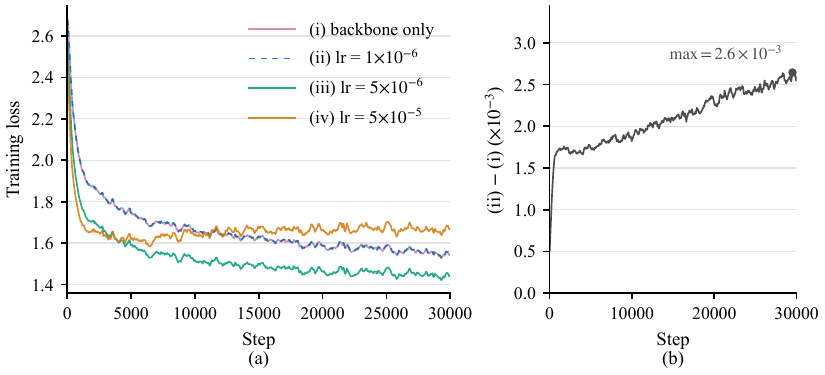}
  \caption{Training loss of incrementally training the n-gram embedding under different
  learning rates, shown over the first $30{,}000$ training steps. (a) Under the single
  learning rate the curve coincides throughout with fine-tuning without the n-gram
  embedding; after the learning rate of the n-gram embedding is increased, the training
  loss drops markedly. (b) Difference between the loss under the single learning rate
  and that of fine-tuning without the n-gram embedding, which remains below
  $2.6\times 10^{-3}$ throughout.}
  \label{fig:pilot-loss}
\end{figure*}

As shown in Figure~\ref{fig:pilot-loss}, with the n-gram learning rate at
$1\times 10^{-6}$ the training loss nearly coincides with that of fine-tuning without the
n-gram embedding: under a single learning rate, the n-gram embedding brings no gain.
We further examine this by measuring the actual contribution of the n-gram embedding to
the forward pass.
Specifically, for each token position $t$ and each injection layer $l$, we compute the
ratio $\frac{\|\mathbf{y}_{l,t}\|}{\|\mathbf{h}_{l,t}\|}$ between the norm of the n-gram
embedding output $\mathbf{y}_{l,t}$ and that of the hidden state $\mathbf{h}_{l,t}$;
if the ratio is tiny, the n-gram embedding produces no effective perturbation and
receives no effective gradient during training.

\begin{table*}[h]
  \centering
  \caption{Contribution of the n-gram embedding at each injection layer, measured at
  step $30{,}000$ under different learning rates of the n-gram embedding.
  $\mathbf{y}_{l,t}$ and $\mathbf{h}_{l,t}$ are the n-gram embedding output and
  the hidden state at layer $l$, position $t$, respectively.}
  \label{tab:gate-stats}
  \begin{tabular}{lcccccc}
    \toprule
   Learning rate & \multicolumn{2}{c}{$1\times 10^{-6}$} & \multicolumn{2}{c}{$5\times 10^{-6}$}& \multicolumn{2}{c}{$5\times 10^{-5}$} \\
    \cmidrule(lr){2-3} \cmidrule(lr){4-5} \cmidrule(lr){6-7}
    Injection position & layer 0 & layer 14 & layer 0 & layer 14& layer 0 & layer 14 \\
    \midrule
    $\frac{\|\mathbf{y}_{l,t}\|}{\|\mathbf{h}_{l,t}\|}$ & 0.050 & 0.009 &0.075&0.012& 0.177 & 0.123 \\
    \bottomrule
  \end{tabular}
\end{table*}

As Table~\ref{tab:gate-stats} shows, at step $30{,}000$ the output of the n-gram
embedding under the single learning rate remains negligible relative to the hidden state
it is added to, and the n-gram embedding table is still close to its initial
state---the forward pass is in effect computed by the backbone alone.
When the learning rate of the n-gram embedding is raised to $5\times 10^{-6}$---the same
ratio to the backbone's learning rate as in Engram
\citep{cheng2026conditional}---curve
(iii) of Figure~\ref{fig:pilot-loss} drops clearly below curve (i), the loss of
fine-tuning without the n-gram embedding.
The loss curves alone, however, cannot tell whether this drop comes from the n-gram
embedding or from the backbone: the ratio at this learning rate is still as small as
0.075 and 0.012 in Table~\ref{tab:gate-stats}, so the drop of curve (iii) is mainly
the drop of the backbone rather than that of the n-gram embedding.
Only at $5\times 10^{-5}$ does the ratio grow markedly (0.177 and 0.123): the
non-updating of the n-gram embedding is alleviated, and the contribution of the n-gram
embedding rises.
We therefore split the backbone and the n-gram embedding into two parameter groups with
different learning rates (Figure~\ref{fig:engram}), forming the dual-learning-rate
scheme: the backbone takes $1\times 10^{-6}$, and the n-gram embedding takes
$5\times 10^{-5}$, fifty times that of the backbone.
All subsequent experiments run under the dual learning rate.

\subsection{Vocabulary size of the n-gram embedding}
\label{sec:method-config}
The vocabulary size decides how many n-grams the n-gram embedding can hold, and it is
fixed before attachment and cannot be adjusted during training.
The vocabulary size of the n-gram embedding involves a trade-off: a larger vocabulary
can store more n-grams, but also requires more data to train; a smaller vocabulary
converges more easily, but different n-grams are then mapped to the same slot, causing
collisions that hurt the expressiveness of the n-gram embedding.
Let the tokenizer vocabulary be of size $|V|$; the number of all n-grams is
$\binom{|V|}{n}$, so using it directly as the vocabulary size of the n-gram embedding is
infeasible.
In joint pretraining, existing works set the vocabulary size of the n-gram embedding to
tens to hundreds of times the tokenizer vocabulary of the backbone
\citep{cheng2026conditional,qwen2026next};
under incremental training, we wish to determine a reasonable size under which
the n-gram embedding is sufficiently trained on limited data while covering
enough domain vocabulary.
We sweep $M\in\{60\mathrm{k},\,152\mathrm{k},\,1520\mathrm{k}\}$.
The sweep runs under the dual learning rate determined in
Section~\ref{sec:method-dual-lr}, with the injection positions fixed at the default given
in Section~\ref{sec:method-setup}.
%
\begin{table*}[h]
  \centering
  \caption{Sweep over the vocabulary size of the n-gram embedding.
  Injection positions fixed; the SFT w/o n-gram embedding row serves as reference.}
  \label{tab:method-vocab}
  \begin{tabular}{lcc}
    \toprule
    Vocabulary size & Telecom (en) & Telecom (zh)\\
    \midrule
    SFT w/o n-gram embedding & 36.59 & 18.81  \\
    \midrule
    $M=60\mathrm{k}$   & 37.86 & 19.09  \\
    $M=152\mathrm{k}$  & 38.70 &  21.66 \\
    $M=1520\mathrm{k}$ & \textbf{41.99} & \textbf{23.18} \\
    \bottomrule
  \end{tabular}
\end{table*}

\begin{figure*}[h]
  \centering
  \includegraphics[width=0.95\linewidth]{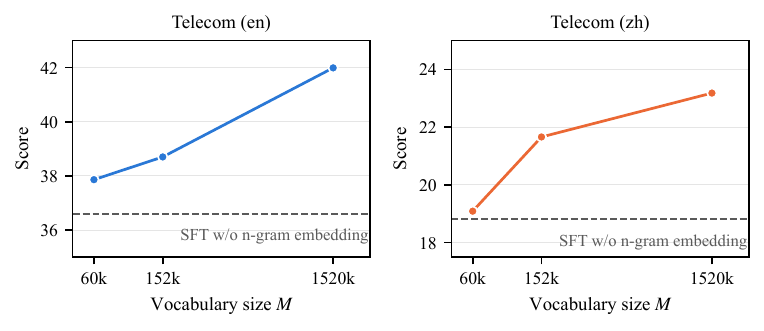}
  \caption{Performance on the two telecom test sets versus the vocabulary size $M$
  of the n-gram embedding (log scale on the horizontal axis). The score rises
  monotonically with $M$ on both test sets, and every configuration surpasses the
  SFT baseline without the n-gram embedding (dashed line).}
  \label{fig:method-vocab}
\end{figure*}
As Table~\ref{tab:method-vocab} and Figure~\ref{fig:method-vocab} show, enlarging the
vocabulary from $60\mathrm{k}$ to $1520\mathrm{k}$ improves performance: a larger vocab size 
lowers hash collisions and raises the resolution of n-gram keys.
The gain, however, does not scale with the parameter cost: a $25\times$ enlargement of
the slot space brings only about 4 points on the two telecom test sets ($+4.13$ on
Telecom (en) and $+4.09$ on Telecom (zh)).
Meanwhile, a larger vocabulary introduces more parameters, which require more data and
training steps to be sufficiently trained.
Balancing the collisions of a small vocabulary against the training cost of a large
one, we select $M=1520\mathrm{k}$, ten times the tokenizer vocabulary of the backbone,
and fix it in the subsequent experiments.

\subsection{Injection position of the n-gram embedding}
\label{sec:method-inject}

The injection position decides at which level of semantics the retrieved n-gram vector is
fused with the backbone:
shallow hidden states stay closer to surface lexical features,
whereas middle-layer hidden states represent higher-level semantics.
In joint pretraining the backbone and the n-gram embedding are optimized simultaneously,
and the model is insensitive to the injection position---the performance gap across
layers is below $1\%$ in \citet{qwen2026next};
under incremental training, in contrast, the backbone parameters update slowly and the
training data are limited, so the injection position may markedly affect performance.

We compare the following injection positions: the embedding layer (layer 0), layer 1,
the middle layer (layer 14), layers $\{0,14\}$, and layers $\{1,14\}$.
Comparing the embedding layer and layer 1 serves to analyze which of the
shallow layers the n-gram embedding fits best;
we further compare the performance of shallow versus middle injection, and whether the
two substitute for or complement each other.
The sweep likewise runs under the dual learning rate, with the vocabulary size fixed at
the $1520\mathrm{k}$ selected in the previous subsection.

\begin{table*}[h]
  \centering
  \caption{Sweep over the injection position of the n-gram embedding.
  Vocabulary size fixed at $1520\mathrm{k}$; the SFT w/o n-gram embedding row serves as
  reference.}
  \label{tab:method-inject}
  \begin{tabular}{lcc}
    \toprule
    Injection position & Telecom (en) & Telecom (zh)  \\
    \midrule
    SFT w/o n-gram embedding & 36.59 & 18.81  \\
    \midrule
    Layer 0 (embedding layer) &  38.95&   21.62\\
    Layer 1                   &  37.24& 21.23\\
    Layer 14 (middle layer)   & 41.48 &  18.92 \\
    Layers $\{0,14\}$         &\textbf{41.99}  &  \textbf{23.18} \\
    Layers $\{1,14\}$         & 41.65 &  20.29\\
    \bottomrule
  \end{tabular}
\end{table*}
As Table~\ref{tab:method-inject} shows, shallow injection and middle injection each
bring a gain over the SFT baseline, and the two single positions each lead on one of
the two test sets: layer 0 scores higher on Telecom (zh), while layer 14 scores higher
on Telecom (en).
Between the two shallow positions, layer 0 outperforms layer 1 on both test sets,
whether injected alone or combined with layer 14.
Combining shallow and middle injection works best---layers $\{0,14\}$ attains the
highest scores on both test sets---indicating that shallow injection and middle
injection complement rather than substitute for each other.
Subsequent experiments therefore take layers $\{0,14\}$ as the injection position.

\section{Experiments}

Section~\ref{sec:method} determined the configuration of the n-gram embedding.
In this section we train the model with that configuration on the
question-answering task in the telecom domain, and evaluate how the n-gram
embedding actually performs under incremental training through the change in
domain performance and general ability after training.
In addition, we analyze how the n-gram embedding takes effect in incremental training
from both qualitative and quantitative perspectives, via a case study and the unloading
experiment.

\paragraph{Setup.}
The training data mixes the telecom question-answering dataset with general
data.
Evaluation covers two sides: the telecom evaluation sets assess the
professional ability of the model, and MMLU and BIG-bench assess its general
ability.
The baselines are the Qwen3-0.6B base model and conventional SFT on the same
data.
Training with the n-gram embedding adopts the configuration determined in
Section~\ref{sec:method}: the dual learning rate ($1\times 10^{-6}$ for the backbone and
$5\times 10^{-5}$ for the n-gram embedding), the vocabulary size
$1520\mathrm{k}$, and the injection positions at layers 0 and 14.

\subsection{Main results}
\label{sec:exp-main}

\begin{table*}[h]
  \centering
  \caption{Incremental training with the n-gram embedding versus the baselines.}
  \label{tab:exp-main}
  \begin{tabular}{lcccc}
    \toprule
     & Telecom (en) & Telecom (zh) & MMLU & BIG-bench \\
    \midrule
    Qwen3-0.6B                      & 31.45 & 12.73 & 48.33 & 41.76 \\
    SFT                     & 36.59 & 18.81 & 48.87 & 45.10 \\
    SFT w/ n-gram embedding & \textbf{41.99} & \textbf{23.18} & 42.23 & 39.80 \\
    \bottomrule
  \end{tabular}
\end{table*}

\paragraph{Results and analysis.}
Incremental training with the n-gram embedding stably surpasses SFT on the domain
tasks.
As shown in Table~\ref{tab:exp-main}, training the n-gram embedding incrementally
exceeds SFT on both domain evaluation sets.
On the general side, the domain gains come at the cost of general ability:
the scores on MMLU and BIG-bench drop below those of both the base model and SFT.

\subsection{Case study}
\label{sec:exp-case}

By Equations~\ref{eq:inject} and~\ref{eq:inject-vec}, the n-gram embedding is injected
into the hidden state
weighted by the gate $\alpha_{l,t}$, and the gate value decides the injection strength
of the n-gram embedding at each token.

\begin{figure*}[h]
  \centering
  \includegraphics{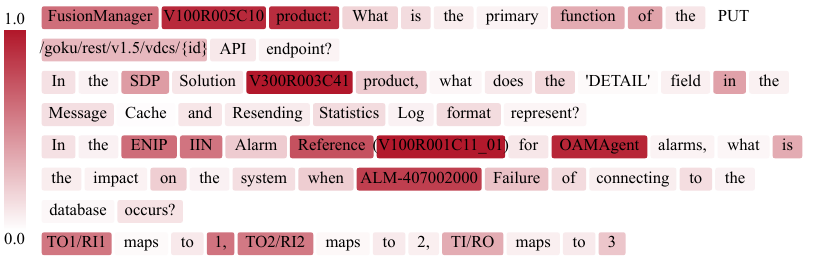}
  \caption{Heatmap of the gate values, averaged over the tokens of each word, after
  incremental training with the n-gram embedding.
  The injection strength of the n-gram embedding is pronounced on domain-relevant
  words.}
  \label{fig:case-gate}
\end{figure*}

To observe the contribution of the n-gram embedding to generation, we take some
examples and visualize the gate values, as shown in Figure~\ref{fig:case-gate}.
Specifically, the text is split into words by spaces, and the gate values of the
tokens within each word are averaged to give a per-word value for the heatmap.

As Figure~\ref{fig:case-gate} shows, the injection strength of the n-gram
embedding is pronounced on domain-relevant words:
the product model number FusionManager V100R005C10, the error code ALM-407002000,
and the interface names TO1/RI1 are all recognized by the n-gram embedding with
high gate values.
By contrast, ordinary digits (2 and 3 in the last row) are not anomalously
activated, owing to the context-awareness of the gate.
This indicates that during incremental training the n-gram embedding learns
domain-relevant words into its vocabulary, and that when needed the gate correctly
recognizes them and successfully intervenes in the hidden-state outputs.

\subsection{Locating where domain knowledge is stored}
\label{sec:exp-unload}

Beyond the qualitative analysis of the case study, we wish to quantitatively locate
where the domain knowledge is stored.
Meanwhile, we want to know whether the performance gain still relies on the training of
the backbone during incremental training.
The additive side-branch structure of the n-gram embedding (Figure~\ref{fig:engram})
allows the two to be told apart: setting the output of the n-gram embedding to zero
leaves only the backbone, with $\theta$ unchanged.
We compare the base model, the model after incremental training with the n-gram
embedding (\textbf{SFT w/ n-gram embedding}), and the model with the n-gram embedding unloaded
(\textbf{n-gram embedding unloaded}), on both the domain and the general tasks.

\begin{table*}[h]
  \centering
  \caption{Unloading experiment: locating where domain knowledge is stored. The two
  n-gram embedding rows share the same set of weights and differ only in whether the n-gram embedding branch
  is enabled at inference.}
  \label{tab:exp-unload}
  \begin{tabular}{lcccc}
    \toprule
    & Telecom (en) & Telecom (zh) & MMLU & BIG-bench \\
    \midrule
    Qwen3-0.6B        & 31.45 & 12.73 & 48.33 & 41.76 \\
    SFT w/ n-gram embedding     &  41.99 & 23.18 & 42.23 & 39.80 \\
    n-gram embedding unloaded   & 27.91 & 13.37 & \textbf{49.70} & \textbf{44.37} \\
    \bottomrule
  \end{tabular}
\end{table*}

\paragraph{Results and analysis.}

As shown in Table~\ref{tab:exp-unload}, unloading the n-gram embedding causes a marked
drop in domain ability---Telecom (en) falls from $41.99$ to $27.91$ and Telecom (zh)
from $23.18$ to $13.37$, both falling back close to the level of the base model;
meanwhile, MMLU and BIG-bench rise instead of falling, exceeding the base model.
This indicates that the domain gain is carried predominantly by the n-gram
embedding---unloading it strips away the gain brought by incremental training---while
the general ability of the backbone remains intact; it also shows that during
incremental training the two did not couple in a way that collapses the backbone when
it generates alone.
In other words, after incremental training the n-gram embedding and the backbone form a
division of labor: domain knowledge is injected into the n-gram embedding, and the
general ability of the backbone is left unaffected.
Compared with plain fine-tuning, where the trained model is a single fixed point, the
branch adds a switch at inference: the same set of weights can serve both
ends---SFT w/ n-gram embedding surpasses the base model on the domain tasks, and
n-gram embedding unloaded surpasses it on the general ones.
The loaded configuration, however, still trails the base model on the general side,
suggesting that the gate does not fully suppress the injection on general inputs;
refining the gate, or introducing an additional sample-level routing on top of it,
would let the model gain domain ability without sacrificing the general one.

\section{Conclusion}
This paper studies transferring the n-gram embedding from the joint-pretraining
scenario to the incremental-training scenario, and extends pretrained models that lack
n-gram embeddings with the n-gram embedding.
By exploring the three factors---the learning rate, the vocabulary size, and the
injection position---we determine the optimal configuration of the n-gram embedding
under incremental training.
On the question-answering task in the telecom domain, the training results show that
incremental training with the n-gram embedding stably improves the domain ability of
the model; meanwhile, the domain ability is trained predominantly into the n-gram
embedding, and the general ability of the backbone is preserved.

\bibliography{ref}

\end{document}